\documentclass[letterpaper,twocolumn,fleqn]{article} 

\usepackage{ist}
\usepackage{array}
\newcolumntype{P}[1]{>{\centering\arraybackslash}p{#1}}
\newcolumntype{M}[1]{>{\centering\arraybackslash}m{#1}}
\usepackage{caption}
\usepackage{bbm}
\usepackage{amssymb}
\usepackage{mathrsfs}
\usepackage{amsmath}
\usepackage[export]{adjustbox}
\usepackage{xcolor}
\title{Self-Supervised Visual Representation Learning on Food Images}
\author{Andrew Peng, Jiangpeng He, Fengqing Zhu \newline\newline 
School of Electrical and Computer Engineering, Purdue University, West Lafayette, Indiana, United States}

\begin{document} 

\maketitle
\thispagestyle{empty} 

\begin{abstract}
Food image analysis is the groundwork for image-based dietary assessment, which is the process of monitoring what kinds of food and how much energy is consumed using captured food or eating scene images. Existing deep learning-based methods learn the visual representation for downstream tasks based on human annotation of each food image. However, most food images in real life are obtained without labels, and data annotation requires plenty of time and human effort, which is not feasible for real-world applications. To make use of the vast amount of unlabeled images, many existing works focus on unsupervised or self-supervised learning of visual representations directly from unlabeled data. However, none of these existing works focus on food images, which is more challenging than general objects due to its high inter-class similarity and intra-class variance.

In this paper, we focus on the implementation and analysis of existing representative self-supervised learning methods on food images. Specifically, we first compare the performance of six selected self-supervised learning models on the Food-101 dataset. Then we analyze the pros and cons of each selected model when training on food data to identify the key factors that can help improve the performance. Finally, we propose several ideas for future work on self-supervised visual representation learning for food images.
\end{abstract}

\section{Introduction}
\label{sec:intro}
Poor diet choices are linked to several health conditions such as cancer, heart diseases, and diabetes, some of the leading preventable causes of death. Additionally, the CDC reports that 9 in 10 Americans consume too much sodium, which may cause high blood pressure, heart disease, and strokes. Furthermore, nearly \$173 billion is spent annually on health care for obesity~\cite{CDC}. However, it is difficult to accurately assess the dietary intake of a person, as traditional methods~\cite{traditional1, traditional2} are based on self-reported information which may include errors due to recall or bias. On the other hand, image-based dietary assessment technologies~\cite{shao2021_ibdasystem} utilize eating occasion images captured by participants to determine their dietary intake. Due to the reduced amount of human input, such technologies can greatly improve the accuracy and reliability of a person's dietary information.

\begin{figure}[t]
    \centering
    \includegraphics[width=1.\linewidth]{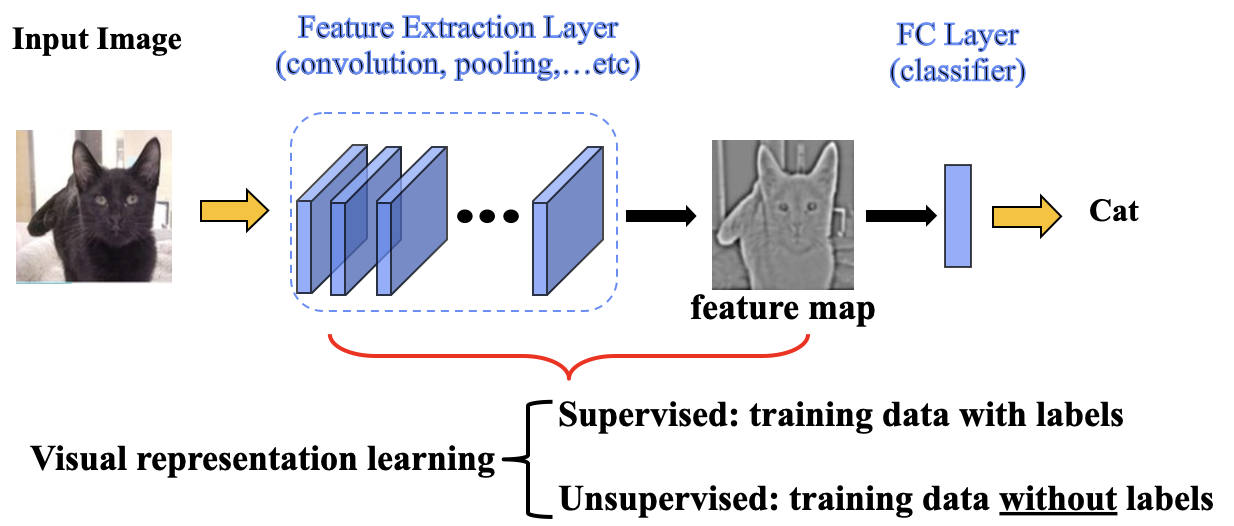}
    \caption{An overview of supervised and unsupervised visual representation learning.}
    \label{fig:unsupervised_image}
\end{figure}

Nowadays, the vast majority of image-based dietary assessment technologies leverage deep learning for food recognition~\cite{mao2020visual, he2022long, pan2022simulating, He_2021_ICCVW}, segmentation~\cite{CNNfood} and portion size estimation~\cite{shao2021towards, he2020multitask, he2021end}. One of the major challenges of existing supervised methods, however, is their requirement for large amounts of annotated training data. Since most food images in real world are captured without labels, an additional step for data annotation is needed, which would be expensive and time-consuming. On the other hand, unsupervised and self-supervised learning models~\cite{unsupervised1, unsupervised2, jigsaw} can learn visual representations directly from unlabeled data to perform downstream tasks. As shown in Fig.~\ref{fig:unsupervised_image}, unsupervised learning trains a feature encoder from unlabeled images to classify the image into a certain category. We will focus on self-supervised learning as it concentrates on downstream tasks while unsupervised learning is more for clustering and dimensionality reduction.
Though numerous deep learning approaches have been developed for self-supervised learning of general tasks, none of them have been tested specifically on food images, which is known to be more challenging due to their intra-class diversity and inter-class similarity~\cite{mao2020visual}.

In this paper, we aim to explore the performance of existing self-supervised learning methods on food images and provide insightful potential directions on improving the performance in future works. We select six state-of-the-art self-supervised learning methods including SimCLR~\cite{SimCLR}, SwAV~\cite{SwAV}, BYOL~\cite{BYOL}, SimSiam~\cite{SimSiam}, MoCo v2~\cite{MoCo, MoCov2}, and DINO~\cite{DINO}, which are representative contrastive based, non-contrastive based and vision transformer based methods, respectively. We evaluate and analyze the self-supervised learning performance on Food-101 dataset~\cite{Food101}, which contains 101 different foods with 1000 images each. Specifically, we first show that self-supervised learning on food images is more challenging by comparing the performance between Food-101 and reported results on ImageNet~\cite{ImageNet}. Then, we analyze the pros and cons of each selected model based on their performance on Food-101 to identify several insights and propose possible future steps for increasing the accuracy and efficiency of self-supervised learning on food images. The contribution of this work can be summarized as the following.
\begin{itemize}
    \item To best of our knowledge, we are the first to systematically study the existing representative self-supervised methods on food images.
    \item We conduct extensive experiments on Food-101 to identify the challenges behind learning on food images compared to general tasks such as ImageNet.
    \item By analyzing the results, we provide insightful directions to potentially improve the performance in future works.
\end{itemize}

\section{Related Work}
\label{sec:relate work}
Self-supervised learning (SSL) and unsupervised learning are two types of models that are trained on completely unlabeled data. Unsupervised learning methods mainly focuses on finding patterns and clustering the data based on similar features. On the other hand, SSL methods attempt to solve tasks by augmenting the unlabeled data, such as rotating the images or taking two different augmentations of the same image.

In this work, we analyze three main categories of SSL models: \textit{Contrastive Based Learning}, \textit{Non-Contrastive Based Learning} and \textit{Vision Transformers (ViTs) Based}. We selected these three categories because of their proven effectiveness on the ImageNet dataset, and because they are three of the most common types of self-supervised image classification models. Furthermore, both Contrastive and Non-contrastive based models that we examine are Siamese models, which are models that compare two augmentations of the same image to learn visual representations. Below we summarize and illustrate each category in detail.

(1) \textbf{Contrastive-based learning} is a common self-supervised learning algorithm that takes two augmentations of the same image, called positive pairs, and maximizes the agreement between the two, while also minimizing the agreement between two augmentations of different images, called negative pairs. One common drawback of contrastive models, however, is their necessity for larger batch sizes, since they require both positive and negative pairs. We selected two most popular methods in this category including SimCLR~\cite{SimCLR} and MoCo v2~\cite{MoCo, MoCov2}, which has a similar structure and comparable performance. Additionally, SimCLR and MoCo v2 are based on similar ideas, with MoCo coming out first and later revised after SimCLR's publish to MoCo v2~\cite{MoCov2}. SimCLR follows the straightforward contrastive learning framework of comparing both positive and negative pairs, with an additional custom optimizer. MoCo instead uses a memory bank to store negative pairs, with MoCo v2 utilizing multi-crop to further increase the performance.

(2) \textbf{Non-contrastive based} learning models also utilize the idea of positive pairs while excluding the negative pairs. They utilize additional techniques to improve performance and also prevent the model from collapsing~\cite{collapse}, which is a common failure in Siamese models when the encoder outputs a constant representation regardless of input. On the other hand, contrastive-based learning models do not collapse since they also contrast negative pairs. For this category, we selected SwAV~\cite{SwAV}, BYOL~\cite{BYOL}, and SimSiam~\cite{SimSiam}, three models that each have their own unique components. SwAV incorporates online clustering within a Siamese model. Similar to MoCo v2, this model also uses multi-crop. Alternatively, BYOL employs two neural networks, called the online and target network, to predict each other's representation of the same image, along with a momentum encoder. Finally, SimSiam~\cite{SimSiam} is a simplistic network that only uses positive pairs and a stop-gradient operation, which prevents certain parts of the model from being updated to prevent collapsing.

(3) \textbf{Vision Transformers}~\cite{ViT} are another type of framework based on the self-attention-based Transfomer~\cite{Transformer}, which is the main method used in Natural Language Processing (NLP). For image tasks, attention-based models have historically under-performed compared to convolutional models such as CNNs and ResNet~\cite{ResNet} due to inefficiencies, but are recently improving in accuracy and computational speed due to modifications on the attention portion of ViTs. One of the main discrepancy between the CNN and ViTs is that Vision Transformers~\cite{ViT} lack certain inductive biases that CNNs are able to produce, such as locality. However, ViTs' accuracy scale up based on the amount of image data, making them ideal for self-supervised image classification where models are typically trained on millions of images. For this category, we selected DINO~\cite{DINO}, a state-of-the-art ViT-based model that also utilizes knowledge distillation~\cite{distillation}. We selected DINO because of its high performance and uniqueness and also we hope to explore other self-supervised learning approaches outside of contrastive and non-contrastive learning to see if they would perform better on food images.


\section{Method}
\label{sec: method}

In this section, we illustrate the selected methods in detail from the perspective of each main category. 
\subsection{Contrastive based Learning}

\begin{figure}[h]
    \centering
    \includegraphics[width=8cm]{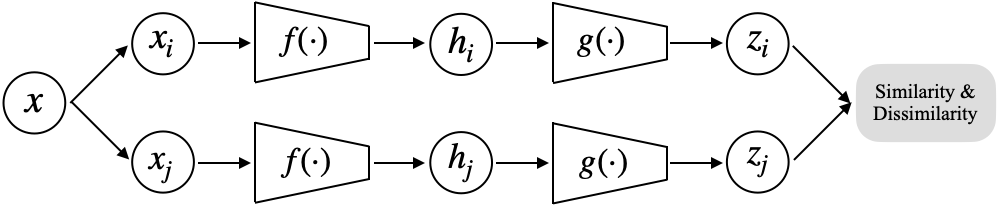}
    \caption{An overview of contrastive learning framework.}
    \label{fig:contrastive_image}
\end{figure}

\textbf{SimCLR and MoCo v2} both exhibit all the features of contrastive learning, using both positive and negative pairs. A brief overview of contrastive learning framework is shown in Fig.~\ref{fig:contrastive_image}, which include four main components: (1) a data augmentation module that randomly generates the two different views of the same image $x$, denoted as $x_i$ and $x_j$, (2) the common ResNet~\cite{ResNet} encoder, represented as $f(\bullet)$, that extracts representation features from the image, (3) a projection head $g(\bullet)$ that calculates contrastive loss, defined as an multilayer perceptron (MLP) with one hidden layer for SimCLR and two hidden layers for MoCo v2 with ReLU~\cite{ReLU}, and (4) a contrastive loss function used to maximize the similarity between positive pairs and minimize the similarity between negative pairs.

SimCLR and MoCo v2 use different contrastive loss functions. SimCLR uses \emph{NT-Xent} (the normalized temperature-scaled cross-entropy loss), defined as in Eq.~\eqref{eq: simclr}. 
\begin{equation}
\begin{aligned}
        \ell_{i,j} = -\log\frac{\exp(sim(z_i, z_j)/\tau)}{\Sigma_{k=1}^{2N}\mathbbm{1}_{[k\neq i]}\exp(sim(z_i, z_k)/\tau)}
    \label{eq: simclr}
\end{aligned}
\end{equation}
where $sim(u,v)=u^Tv/|u||v|$, $N$ is the batch size, $\mathbbm{1}_{[k\neq i]} \in {0, 1}$ is an indicator function evaluating to $1$ iff $k\neq i$ and $\tau$ denotes a temperature hyper-parameter. 
MoCo v2, on the other hand, uses InfoNCE, defined as in Eq.~\eqref{eq: mocov2}. 
\begin{equation}
\begin{aligned}
        \mathcal{L}_q = -\log \frac{\exp(q \cdot k_i)/ \tau}{\Sigma_{i=0}^k \exp(q\cdot k_+)/ \tau}
    \label{eq: mocov2}
\end{aligned}
\end{equation}
where $q$ is the query, $k_+$ is the positive key, $k_i$ are the other keys, and $\tau$ also denotes a temperature hyper-parameter. These equations represent the final step of Fig.~\ref{fig:contrastive_image}, and both of these functions aim to maximize the similarity between positive pairs and minimize the similarity between negative pairs. In this work, we select these two methods to represent the contrastive-based models and further evaluate the performance on food images, which can be more challenging due to the intra-class diversity and inter-class similarity.

\subsection{Non-Contrastive based Learning}
\begin{figure}[h]
    \centering
    \includegraphics[width=8cm]{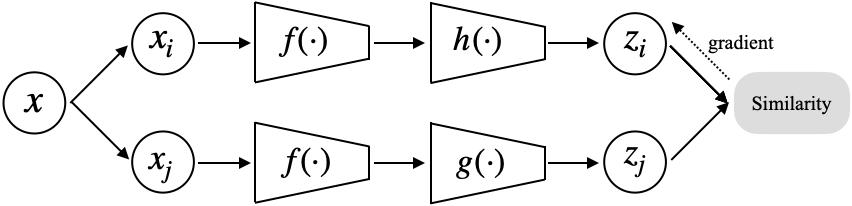}
    \caption{An overview of non-contrastive learning framework.}
    \label{fig:non-contrastive_image}
\end{figure}

\textbf{SwAV, BYOL, and SimSiam} are non-contrastive based models as they do not incorporate negative pairs, but they are all Siamese models because they compare positive pairs. Additionally, since all three models are structured the same conceptually, we will examine SimSiam in detail as it is more representative compared to BYOL and SwAV, which both have additional unique features. SimSiam also demonstrates that they can learn visual representations with only a stop-gradient operation. In Fig.~\ref{fig:non-contrastive_image}, an overview of non-contrastive learning is shown, which demonstrates the fundamental framework of our three models. Similar to SimCLR, $f(\bullet)$ represents an encoder while $h(\bullet)$ denotes the MLP head.

SimSiam utilizes a simplistic loss function defined as in Eq.~\eqref{eq: simsiam}:
\begin{equation}
        \mathcal{L}= \frac{1}{2}\mathcal{D}(a_1, \text{stopgrad}(b_2))+\frac{1}{2}\mathcal{D}(a_2, \text{stopgrad}(b_1))
    \label{eq: simsiam}
\end{equation}
where $a_1=h(f(x_1))$ and $b_2=f(x_2)$ and $\mathcal{D}(a_1, b_2)$ represents the negative cosine similarity: $-\frac{a_1}{\| a_1 \|_2} \cdot \frac{b_2}{\| b_2 \|_2}$, where $\| \cdot \|_2$ is the $l_2$-normalized vector. Since both SwAV and BYOL are different from SimSiam, we will consider their unique features, represented by $g(\bullet)$. SwAV utilizes online clustering with Sinkhorn-Knopp transform~\cite{Sinkhorn} and BYOL directly predicts the output of one view from another view using a momentum encoder. Both these methods are used to prevent collapsing and improve accuracy. Although non-contrastive methods do not utilize negative pairs, they are all still Siamese networks. Therefore, we evaluate their performance to see if visual representations of food images will be learned well by these methods.



\subsection{Vision Transformer-based Learning}
\begin{figure}[h]
    \centering
    \includegraphics[width=8cm]{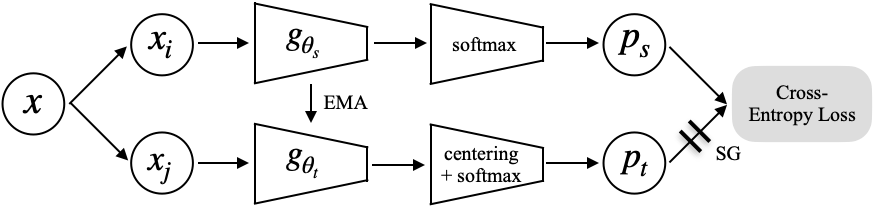}
    \caption{An overview of the DINO model.}
    \label{fig:dino_image}
\end{figure}

\textbf{DINO} is the representative model we selected in Vision Transformer based category. It is a self-supervised learning approach that utilizes \emph{self-distillation with no labels}. In addition to SSL, it utilizes knowledge distillation~\cite{distillation}, which involves training a student network's probability distribution based on an input image to match the output of a teacher network. By maximizing the similarity between their predictions and propagating the information to update the networks, the model is able to learn visual representations of different images. An overview of DINO is shown in Fig.~\ref{fig:dino_image}. A positive pair is passed into the two networks represented by $g$, which are composed of a backbone ViT and a projection MLP head similar to the one used in SwAV. The loss function is defined as $\min H(P_t(x), P_s(x))$, which takes the cross-entropy loss of probability distributions of the teacher and student network. $H(a,b) = -a\log b$ and $P_s$ is defined as in Eq.~\eqref{eq: dino}:
\begin{equation}
        P_s = \frac{\exp (g_{\theta_s}(x)/\tau_s)}{\Sigma_{k=1}^K \exp (g_{\theta_s}(x)/\tau_s)}
    \label{eq: dino}
\end{equation}
with $\tau_s$ as a hyperparameter. Additionally, a stop-gradient operator (SG) is applied to propagate gradients only through the student, while the teacher parameters are updated with an exponential moving average (EMA), defined by the formula $\theta_t \leftarrow \lambda \theta_t + (1-\lambda)\theta_s$, where $\theta_t$ and $\theta_s$ are parameters, and $\lambda$ follows a cosine schedule from $0.996$ to $1$. We chose this Vision Transformer-based model because it was fundamentally different from the other methods while also achieving high performance on the ImageNet dataset. Additionally, DINO also uses knowledge distillation, which is another technique we hope to explore on food images.

\section{Experiments}
\label{sec: exp}
In this section, we first evaluate the selected six self-supervised method by comparing the performance on both general object dataset and food image dataset. Then, we specifically analyze the results on food data and summarize the pros and cons of each selected model. Finally, we provide insights as future work to further improve the performance on food images.

\subsection{Experimental Setup}
\textbf{Datasets:} we used the Food-101 dataset, which includes $101$ different food classes with $1,000$ images each, providing a total of $101,000$ food images. Each food class is further divided into $750$ training images with $250$ test images. Additionally, some of the images are purposely uncleaned with a certain amount of noise, such as intense colors and mislabeled images. We selected this dataset as it is one of the most well-known food datasets used for various downstream tasks. 

While focusing on food images, we also leverage ImageNet dataset containing images of general objects as a reference compared to Food-101. The ImageNet Large Scale Visual Recognition Challenge (ILSVRC) 2012-2017 contains $1000$ image classes and over $1.2$ million images, $50,000$ validation images and $100,000$ test images, which is commonly used as a benchmark to evaluate the model performance especially on image classification tasks. 

\begin{figure*}[t]
    \centering
    \includegraphics[width=1. \linewidth]{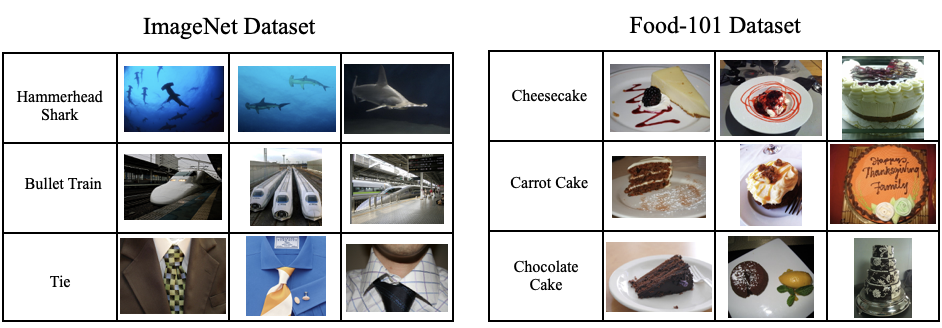}
    \caption{Intra-class diversity and inter-class similarity in Food-101 compared to ImageNet.}
    \label{fig:comparison}
\end{figure*}

As shown in Fig.~\ref{fig:comparison}, the images in Food-101 is of higher intra-class diversity and inter-class similarity compared to ImageNet datasets, making it more challenging to learn the visual representation from unlabeled data.

\textbf{Evaluation metric:} we use the widely applied \textit{Linear Evaluation} as the evaluation metric, which trains a supervised linear classifier on frozen features learned by self-supervised visual representation learning. Specifically, a fully-connected layer followed by softmax is trained on the test images, and the gradients are not propagated back to the frozen features which ensure the feature extractor does not learn anything from supervised labels.

\textbf{Implementation details:} In Table~\ref{Implementation}, we summarize the implementation details of each selected model. We selected 100 epochs for all the models and ResNet-50 for contrastive and non-contrastive-based models. Additionally, we chose reasonable batch sizes of 128/256 for the contrastive and non-contrastive models, which were the maximum allowed by our computational resources. For DINO, we chose 64 batch size due to it being a ViT-based model, which claims to require less batch size.

\begin{table}
\begin{adjustbox}{width=\linewidth}
\centering
\begin{tabular}{ p{1.2cm}||p{1cm}|p{1cm}|p{1cm}|p{1.1cm}|p{1.2cm}|p{1cm}  }
\hline
&SimCLR&SwAV&BYOL&SimSiam&MoCo v2&DINO\\
\hline
Batch Size&256&256&128&128&256&64\\
\hline
Epochs&100&100&100&100&100&100\\
\hline
Backbone&ResNet-50&ResNet-50&ResNet-50&ResNet-50&ResNet-50&ViT-S\\
\hline
Optimizer&LARS&SGD&SGD&SGD&SGD&AdamW\\
\hline
\end{tabular}
\end{adjustbox}
\caption{Implementation Details}
\label{Implementation}
\end{table}

\subsection{Results on Food-101}
\label{subsec: model comparison}
The experimental results of six selected self-supervised methods on Food-101 are summarized in Table~\ref{Food101}. We include the top-1 linear evaluation accuracy (\%) along with training time and how much memory the model parameters use to show that our trained models require a similar amount of computational resources. Only DINO takes up significantly more memory, which is due to ViTs requiring storing more memory after training. From the results, we observe that DINO performed the best while BYOL and SimSiam performed the worst. We expected slightly lower accuracy in SimSiam because it had no unique method to improve accuracy, but BYOL's lower accuracy was unexpected.  Additionally, we notice that the other three models, SimCLR, SwAV, and MoCo v2, have similar accuracy, showing that each model's unique methods increased their accuracy. The performance difference between the best and worst model is approximately 18.5\%, which is quite significant.

\begin{table}
\begin{adjustbox}{width=\linewidth}
\noindent\begin{tabular}{ p{1.2cm}||p{1cm}|p{1cm}|p{1cm}|p{1.1cm}|p{1.2cm}|p{1cm}  }
\hline
&SimCLR&SwAV&BYOL&SimSiam&MoCo v2&DINO\\
\hline
Accuracy (\%)&51.0&54.7&47.7&44.5&53.9&61.4\\
\hline
Training Time&2 days&2 days&3 days&2 days&3 days&2 days\\
\hline
Memory Size&107M&217M&283M&292M&305M&672M\\
\hline
\end{tabular}
\end{adjustbox}
\captionof{table}{Experimental results on Food-101}
\label{Food101}
\end{table}

\subsection{Comparison Between Food-101 and ImageNet}
\begin{figure}[h]
    \centering
    \includegraphics[width=1. \linewidth]{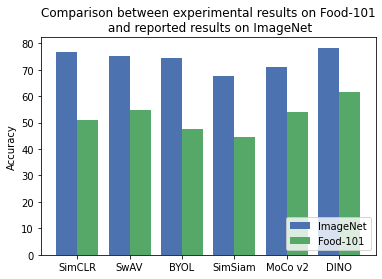}
    \caption{Linear evaluation results on ImageNet and Food-101.}
    \label{fig:bar_plot}
\end{figure}

In Fig.~\ref{fig:bar_plot}, we included the results on Food-101 side-by-side with the results on the ImageNet dataset. These results were obtained using the same number of epochs and backbone encoder.

Our experimental results on Food-101 are much lower than the results on ImageNet. This observation also shows that the visual representation on food images can be more challenging than general objects in real life due to higher intra-class diversity and inter-class similarity, resulting in the lower accuracy. Another reason for the lower accuracy could be that the Food-101 contains fewer images than ImageNet, which hurts the performance of existing self-supervised models as they rely on massive amounts of data for training. 

We also notice that BYOL performs worse than expected with the lowest performance on Food-101 out of all of the selected models, despite it achieving a high accuracy on ImageNet. One of the possible reason is that the BYOL reuiqres larger batch size where it achieves 74.3\% accuracy after training on 512 TPUs with 4096 batch size, while our model was only trained on 128 batch size with ResNet-50. Therefore, we speculate that BYOL and other models could scale up in accuracy with larger batch size and computation resources. To prove our assumption, we ran intermediate experiments using SimCLR as shown in Table~\ref{table:simclr}. The increased batch size results in a notable increase in accuracy even without increasing the number of epochs or changing the backbone network. However, the computation resource is one of the major constrains in deep learning especially for real world applications.   

\begin{table}
\centering
\begin{tabular}{ c||c|c|c }
\hline
Batch Size&64&128&256\\
\hline
Accuracy&41.8&48.0&51.0\\
\hline
\end{tabular}
\caption{SimCLR accuracy on Food-101 with various batch sizes.}
\label{table:simclr}
\end{table}

Additionally, we noticed that DINO performed much better than all the contrastive models. One possible reason is that ViTs perform better than ResNet as the backbone for visual representation learning. This could possibly be due to the fact that self-supervised learning uses a large amount of training data, which benefits ViTs more significantly due to their unique attention modules. Another possible reason is that the lower batch sizes hurt more performance of contrastive learning models, as they require both positive and negative pairs for training. Non-contrastive models are also impacted by lower batch sizes, although to a lesser extent. Therefore, we have demonstrated that batch sizes and, in general, computational resources are more impactful for contrastive and non-contrastive models, while ViTs do not depend as much on batch sizes when compared with contrastive and non-contrastive models.

Finally, we compare the performance of four Siamese models: SimCLR, MoCo v2, SwAV, and SimSiam. We exclude BYOL due to its lower-than-expected accuracy from our comparison. Firstly, SimSiam has a lower accuracy than the other three models, which all have very similar performance, although SwAV and MoCo v2 perform slightly better as they both adopted some ideas from SimCLR. This shows that the unique features in SimCLR, MoCo v2, and SwAV improved their accuracy compared to SimSiam's stop-gradient operation. Furthermore, both SwAV and MoCo v2 add an extra layer of complexity with their unique features, which are online clustering and momentum encoders, respectively. Through this comparison, we see that contrastive and non-contrastive based models perform similarly. Since both models achieved similar high performances, we can conclude that the Siamese learning framework is efficient in learning visual representations. 

\section{Insightful Directions for Future Work}
\label{sec:future works}
Based on our experiments and analysis, we proposed three ideas on how to improve accuracy in the future:
\begin{itemize}
    \item \textbf{Fine-Tuning or Transfer Learning.} This method involves training a model on a large dataset, for example ImageNet, and then transferring on Food-101. This approach could help resolve the issue of an insufficient amount of training data since the model will have learned visual representations on a larger dataset. Therefore, when fine-tuning on a smaller dataset, the model will not require a large batch size to learn visual representations from scratch.
    \item \textbf{Larger Computational Resources.} As already shown in Table~\ref{table:simclr}, accuracy scales up with larger batch sizes and more training epochs. Therefore, we would expect better performance if the models are trained with larger computation resources.
    \item \textbf{Ensemble of Models.} We propose that combining certain models could improve accuracy. We observed that each method category had its own unique techniques which improved their accuracy, so combining some of the methods together may be a potential solution. For example, we researched pre-text tasks, which are unsupervised image-based problems solved to learn the visual representation of an image, such as colorizing a black-and-white image. We will be examining the rotation pre-text task, which predicts if an image is rotated $0^\circ$, $90^\circ$, $180^\circ$, and $270^\circ$, implemented by the model RotNet~\cite{RotNet}. We propose that combining SimCLR with RotNet, for example, could further improve the accuracy because it learns more visual representations.
\end{itemize}

\section{Conclusion}
\label{sec:conclusion}
Overall, we explored the performance of 6 state-of-the-art models from 3 main categories on food images, specifically the Food-101 dataset. Our experimental results show that visual representation learning is more challenging on food images by comparing performance on ImageNet and Food-101. Additionally, all three categories of models show promising results on food data. The experimental results suggest that ViT models are worth exploring further for self-supervised image tasks, but contrastive and non-contrastive models should still be considered when working on self-supervised classification tasks. Finally, based on our analysis, we propose that there is also potential for transfer learning or combining models to help improve accuracy.

\bibliographystyle{IEEEtran}
\bibliography{ref}

\begin{biography}
Andrew Peng is currently a senior at Henry M. Gunn High School in Palo Alto, CA. He conducted research over the summer and throughout 2022–2023 at VIPER Lab with the School of Electrical and Computer Engineering in Purdue University, West Lafayette, IN. His current research interests lie in Image Classification and Signal Processing using deep neural networks. \newline 

Jiangpeng He received his Ph.D. degree in Electrical and Electronic Engineering from Purdue University in August 2022. He is currently a postdoc research assistant at the School of Electrical and Computer Engineering, Purdue University, West Lafayette, IN, USA. His research interests include image processing, computer vision, image-based dietary assessment and deep learning.\newline

Fengqing Zhu is an Associate Professor of Electrical and Computer Engineering at Purdue University, West Lafayette, Indiana. Dr. Zhu received the B.S.E.E. (with highest distinction), M.S. and Ph.D. degrees in Electrical and Computer Engineering from Purdue University in 2004, 2006 and 2011, respectively. Her research interests include image processing and analysis, video compression and computer vision. Prior to joining Purdue in 2015, she was a Staff Researcher at Futurewei Technologies (USA).
\end{biography}

\end{document}